\documentclass[final]{cvpr}

\usepackage{times}
\usepackage{epsfig}
\usepackage{graphicx}
\usepackage{amsmath}
\usepackage{amssymb}
\usepackage{caption}
\usepackage{booktabs}
\usepackage{color}
\usepackage{longtable}
\usepackage{subfig}
\usepackage{tabularx}
\usepackage{boxhandler}
\usepackage{adjustbox}
\usepackage{multirow}
\usepackage{verbatim}
\usepackage{amssymb}
\usepackage{pifont}
\newcommand{\xmark}{\ding{55}}%
\usepackage[justification=centering]{caption}
\usepackage{lipsum}
\newcommand\blfootnote[1]{%
  \begingroup
  \renewcommand\thefootnote{}\footnote{#1}%
  \addtocounter{footnote}{-1}%
  \endgroup
}

\def\cvprPaperID{87} 
\def\confYear{CVPR 2021}

\begin{document}

\title{Noise Conditional Flow Model for Learning the Super-Resolution Space}
\author{Younggeun~Kim\textsuperscript{\rm 1, 3}\qquad~Donghee~Son\textsuperscript{\rm 2, 3} \\
		{\textsuperscript{\rm 1} Seoul National University, Seoul, Republic of Korea
		}  \\
		{\textsuperscript{\rm 2}Lomin Inc., Seoul, Republic of Korea} \\
		{\textsuperscript{\rm 3}Deepest, Seoul, Republic of Korea } \\
		\small{\texttt{eyfydsyd97@snu.ac.kr}} \qquad
		\small{\texttt{dh.son@lomin.ai}}
	}

\maketitle

\begin{abstract}
   Fundamentally, super-resolution is ill-posed problem because a low-resolution image can be obtained from many high-resolution images. Recent studies for super-resolution cannot create diverse super-resolution images. Although SRFlow tried to account for ill-posed nature of the super-resolution by predicting multiple high-resolution images given a low-resolution image, there is room to improve the diversity and visual quality. In this paper, we propose Noise Conditional flow model for Super-Resolution, NCSR, which increases the visual quality and diversity of images through noise conditional layer.
   To learn more diverse data distribution, we add noise to training data. However, low-quality images are resulted from adding noise. We propose the noise conditional layer to overcome this phenomenon. The noise conditional layer makes our model generate more diverse images with higher visual quality than other works. Furthermore, we show that this layer can overcome data distribution mismatch, a problem that arises in normalizing flow models. With these benefits, NCSR outperforms baseline in diversity and visual quality and achieves better visual quality than traditional GAN-based models. We also get outperformed scores at NTIRE 2021 challenge~\cite{NTIRE2021SRSpace}.

\end{abstract}
\blfootnote{This work was done as a project of Deepest}

\section{Introduction}

\begin{figure}
\scriptsize
	\begin{center}
	\includegraphics[width=0.96\linewidth, height=0.7\linewidth]{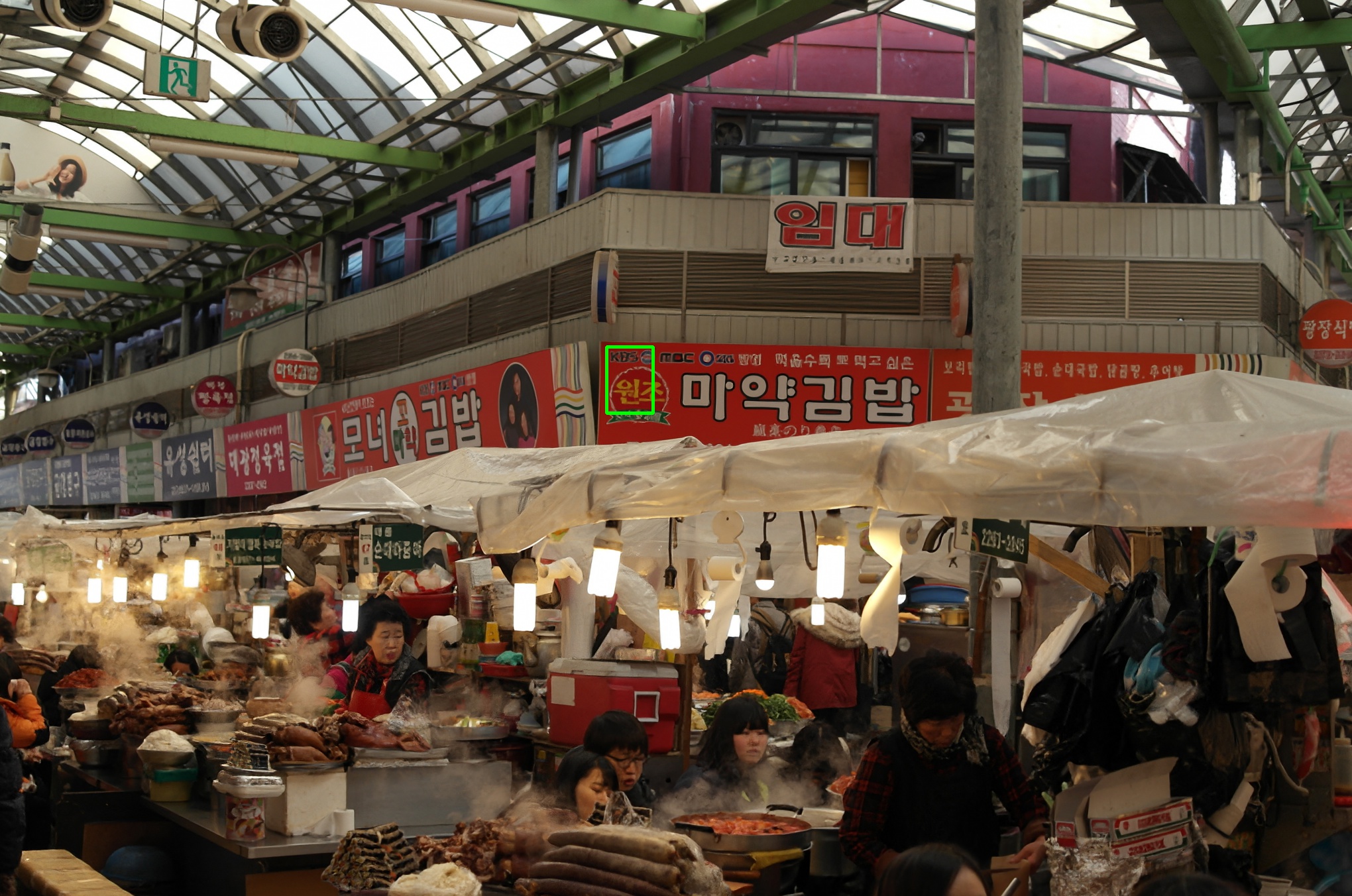}	 \\
		\renewcommand{\tabcolsep}{0.1mm}
		\begin{tabular}{ccccc}
		\includegraphics[width=0.19\linewidth]{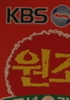} &
		\includegraphics[width=0.19\linewidth]{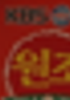} &
		\includegraphics[width=0.19\linewidth]{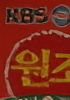} &
		\includegraphics[width=0.19\linewidth]{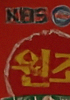} &		
		\includegraphics[width=0.19\linewidth]{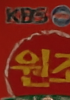} \\
		{HR} & {Bicubic} & {ESRGAN} & {SRFlow} & \textcolor{red}{\textbf NCSR (Ours)} \\
	    \vspace{-0.1cm}
	\end{tabular}
	
    \captionsetup{justification=raggedright,singlelinecheck=false}
    \vspace{-0.2cm}
	\caption{$\times$4 Super-Resolution result of our model on "0831" from DIV2K validation set compared with other baseline models.}
	\end{center}

	\vspace{-1cm}
	\label{fig:front}
\end{figure}

Single image super-resolution is a computer vision task to reconstruct a high-resolution image from its low-resolution image. Super-resolution is an important problem in computer vision due to its various applications including surveillance~\cite{zou2011very}, medical imaging~\cite{chen2018efficient, shi2013cardiac}, astronomical imaging~\cite{puschmann2005super, li2018super} and object detection~\cite{noh2019better}.

With the development of deep learning in computer vision~\cite{goodfellow2014generative, he2016deep, hu2018squeeze, huang2017densely, radford2015unsupervised}, super-resolution methods based on deep learning~\cite{kim2016accurate, ledig2017photo, lim2017enhanced, wang2018esrgan, zhang2018image, zhang2018residual} improve performance significantly. Models~\cite{kim2016accurate, lim2017enhanced, zhang2018image, zhang2018residual} trained with $L_1$ or $L_2$ loss achieve high PSNR performance. Likewise, models~\cite{johnson2016perceptual, ledig2017photo, wang2018esrgan} trained with adversarial loss or perceptual loss accomplish high visual quality performance. 

Most super-resolution methods based on deep learning take a low-resolution image as an input, then output a high-resolution image. However, super-resolution is an ill-posed problem. That is to say, one low-resolution image can be mapped from multiple high-resolution images. SRFlow~\cite{lugmayr2020srflow} proposed the method that can predict multiple high-resolution images for a given low-resolution image by learning the super-resolution space. ~\cite{lugmayr2020srflow} utilize normalizing flow to learn super-resolution space.

SRFlow\cite{lugmayr2020srflow} produces a variety of results other than a deterministic super-resolution output, but there are possibilities to improve performance. Their diversity comes from training with negative log-likelihood, not $L_{1}$ or $L_{2}$ loss. SRFlow~\cite{lugmayr2020srflow} outputs more diverse super-resolution images with better visual quality than not only models that give the deterministic result such as GAN based but also a model that simply creates diversity through noise~\cite{Rakotonirina_2020}. However, in addition to training with negative log-likelihood, there are more ways to increase diversity. 

In this paper, we propose a model that produces results with more advanced diversity and better visual quality than SRFlow. Our method increases diversity by adding noise. However, simply adding noise to input images generates low-quality images with artifacts. To deal with this problem, we propose a structure called noise conditional layer, which results in superior results in both metrics over SRFlow~\cite{lugmayr2020srflow}.
We also analyze that these improvements come from resolving data distribution mismatch that exists in other flow models such as ~\cite{kingma2018glow}. Existing flow-based models aim to map complex data $x$ from simple data $z$, but when the two manifold dimensions are not the same, flow models are not trained smoothly. Therefore, these distribution dimension mismatches should be addressed. This was addressed in SoftFlow\cite{kim2020softflow} and we applied similar ideas to super-resolution tasks. Our contribution is as follows.

\begin{enumerate}
\item We propose a method that can improve performance on learning the super-resolution space using flow model through adding noise and noise conditional layer.
\item  Our method improve the performance of the diversity by adding noise to the training data to expand the data distribution

\item Our method solves the performance degradation caused by the mismatch of the data distribution of SR model using normalizing flow.

\end{enumerate}


\section{Related Works}
\subsection{Single Image Super-Resolution}
As deep learning-based methods~\cite{goodfellow2014generative, he2016deep, hu2018squeeze, huang2017densely, radford2015unsupervised} provide significant performance improvement, many single image super-resolution methods~\cite{kim2016accurate, lim2017enhanced, zhang2018image, zhang2018residual} based on deep learning are proposed. Dong \etal ~\cite{dong2015image} proposed the first super-resolution model based on deep learning. Dong \etal ~\cite{dong2015image} propose the model, which use three convolution layers, trained with $L_{2}$ loss. After that, many methods~\cite{kim2016accurate, lim2017enhanced, zhang2018image, zhang2018residual} which optimized with $L_{1}$ or $L_{2}$ loss are proposed. Although these models show performance improvement in terms of PSNR, some of their predictions are blurry. To deal with this problem, super-resolution models~\cite{ledig2017photo, wang2018esrgan} which use adversarial loss or perceptual loss are proposed. However, these works predict a reconstructed high-resolution image for a given low-resolution image.

\subsection{Normalizing Flow}
Flow-based model, originally introduced in~\cite{dinh2014nice}, proposed deep learning framework for modeling complex high-dimensional density. Flow-based models have made many advances to map accurate complex distributions from simple distribution.
Several approaches such as~\cite{dinh2014nice, dinh2016density, kingma2018glow} use invertible networks to map complex distributions from simple distributions (ex. Gaussian).~\cite{grathwohl2018ffjord} uses a continuous-time invertible generative model with unbiased density estimation and one-pass sampling. Besides,~\cite{kim2020softflow} aims to estimate the conditional distribution of perturbed input data instead of learning the data distribution directly to solve the discrepancy problem of a dimension of data distribution. 
Recently flow-based models are gaining popularity in the field of image generation~\cite{dinh2016density, kingma2018glow}. Moreover,~\cite{ardizzone2019guided, winkler2019learning} present a conditional image generation method based on the Glow architecture.~\cite{winkler2019learning} deals with SR tasks but does not produce influential results compared to GAN-based models. 
For the first time, SRFlow~\cite{lugmayr2020srflow} propose a flow-based super-resolution model which outperforms GAN-based models.
SRFlow~\cite{lugmayr2020srflow} is trained with the negative log-likelihood loss only. By using negative log-likelihood loss, SRFlow~\cite{lugmayr2020srflow} solves the deterministic output problem posed by the previous super-resolution works and learns to generate diverse photo-realistic super-resolution images.
Our method following Glow architecture~\cite{kingma2018glow} along SRFlow, solves the data distribution problem like~\cite{kim2020softflow} and generates super-resolution images of better visual quality and more diverse outputs.

\section{Method}
Our main goal is to learn super-resolution space. In other words, we aim to generate diverse super-resolution images with high visual quality for a given low-resolution image. In this section, we introduce our proposed method. Firstly, we will briefly address the background to understand our model. Next, we will discuss how to improve diversity. Finally, we explain the noise conditional layer which improves image visual quality and diversity by solving mismatch in the distribution of data.

\subsection{Background}
Flow-based model is one of the effective methods for predicting the complex distribution of real data. These models aim to convert from a simple (ex. Gaussian) distribution to a complex (ex. Real-world) using a series of invertible functions. These properties make complex data $x$ can be always reconstructed from $z$, which is the latent vector. These flow-based generative models are defined as: $$z \sim p_{z}(z)$$ $$x=g(z),\ z=f(x)$$ $$f(x) = f_{n}\circ f_{n-1}\circ \cdots \circ f_{1}(z)$$
In this case, $z$ is the latent variable, $f$ and $g$ are invertible to each other, $z=f(x)=g^{-1}(x)$. Moreover, the flow model $f$ consists of invertible transformation, which maps dataset $x$ to Gaussian latent variable $z$ and each $f_{i}$ has a tractable inverse and a tractable Jacobian determinant. The series of invertible transformations is called a normalizing flow, and the advantage of this normalizing flow is that probability density $p_{x}$ can be written as follows by applying the change of variable formula:
$$p_{x}(x|\theta) = p_{z}(f_{\theta}(x))|\det{\frac{\partial{f_{\theta}}}{\partial{x}}}| $$
It allows network to be trained through the following objective functions.
$$-\log{p_{x}(x|\theta)}=-\log{p_{z}(f_{\theta}(x))}- \log{|\det{\frac{\partial{f_{\theta}}}{\partial{x}}}|}$$

Model $f$ is trained by directly minimizing negative log-likelihood.
These training methods prevent the output of the model from being deterministic. This guarantees the diversity of the output.

SRFlow~\cite{lugmayr2020srflow} is a normalizing flow-based super-resolution method that, given a low-resolution image, can learn a super resolution conditional distribution for that image. They utilized the basic Glow architecture and modified the existing flow step to create the conditional flow step. The conditional flow step consists of  Actnorm, 1$\times$1 convolution, Affine injector, and Conditional affine coupling. The parts that learn low-resolution images conditionally are affine injector and conditional affine coupling.\\
Affine injector  is as follow:
$$h^{n+1} = exp(f^{n}_{\theta,s}(u))\cdot h^{n} + f_{\theta,b}^{n}
(u)$$
which, $f^{n}_{\theta,s}$ and $f^{n}_{\theta,b}$ can be any network and $u = g_{\theta}(x)$ is low resolution image encoding , where $g_{\theta}$ is encoder network.\\
Conditional affine coupling is as follow:
$$h_{A}^{n+1} = h_{A}^{n}$$  $$ h_{B}^{n+1} =  exp(f^{n}_{\theta,s}(h_{A}^{n};u))\cdot h_{B}^{n} + f_{\theta,b}^{n}(h_{A}^{n};u)$$
Similar to affine injector, $f^{n}_{\theta,s}$ and $f^{n}_{\theta,b}$ can be any network and $u = g_{\theta}(x)$ is low-resolution image encoding. Moreover, $h^{n}$ = ($h^{n}_{A}$ , $h^{n}_{B}$ ) is a partition in the channel dimension.
\label{background}

\subsection{How to improve Diversity}
Learning the super resolution space  was already addressed in~\cite{lugmayr2020srflow}.~\cite{lugmayr2020srflow} was trained with negative log-likelihood, which does not use the loss as $L_1$ loss. We use~\cite{lugmayr2020srflow} which obtained a meaningful diversity as baseline.

 \begin{figure}[t]
\begin{center}
   \includegraphics[width=1.0\linewidth]{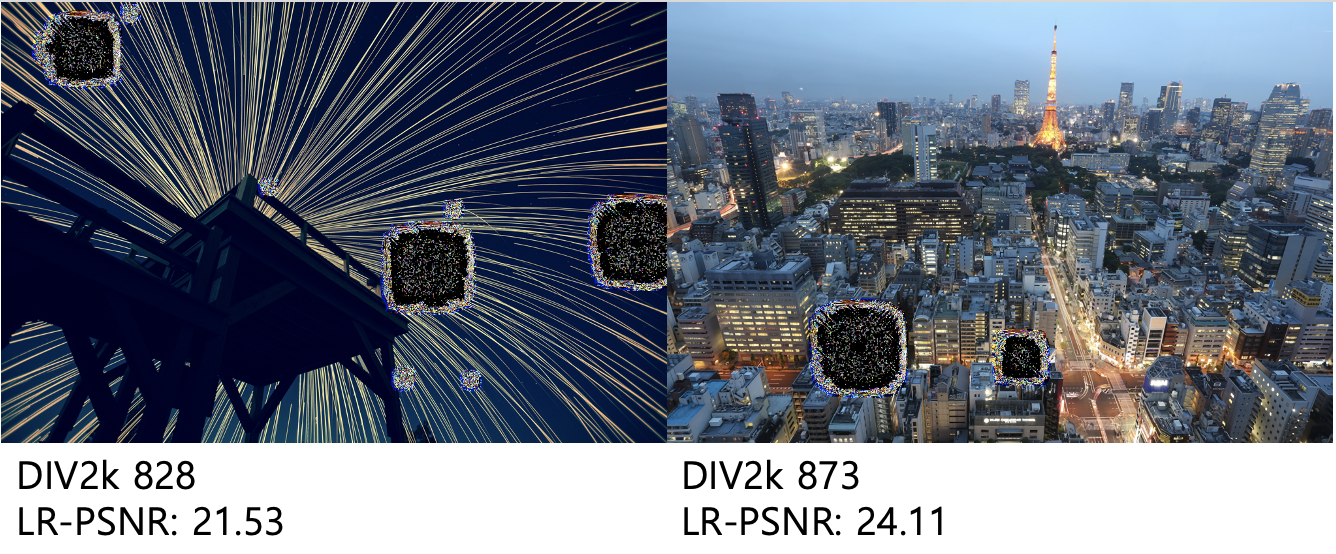}
\end{center}
\vspace{-0.5cm}
\captionsetup{justification=raggedright,singlelinecheck=false}
   \caption{These images are samples of super-resolution results of DIV2k validation set from a model trained only with injecting noise, without noise conditional layer.}
\label{fig:only_noise}
\vspace{-0.3cm}
\end{figure}

To achieve a higher diversity score, we started from the fact that flow-based models aim to match the distribution of simple data $z$ with complex data $x$.
\begin{enumerate}
\item By varying the distribution of complex data $x$, i.e. high-resolution images, the distribution of super-resolution outputs mapped from data $z$ will also vary during the inference process.
\item Among the methods to vary the distribution of this data $x$, we use noise injection.  $$x^+ = x + noise$$  $$f^{-1}(x^+|y)=z $$where, $x$ is high resolution image, $f$ is invertible model that maps $z$ to $x$ and $y$ is low resolution image. However, while the method of varying the distribution by adding noise to the high-resolution image has increased the diversity significantly, this is not the exact LR image conditional training. These training methods generate super-resolution images with severe noise when generating images from simple distribution $z$.
\item  Therefore, for exact LR image conditional training, noise injected into the high-resolution image was resized to the size of LR image and added to the LR image for training.
$$x^+ = x + noise$$ $$y^+ = y+ noise^{-}$$$$ f^{-1}(x^{+}|y^{+})=z$$where, $noise^{-}$ is a vector whose noise is resized in the same size as $y$\end{enumerate}
This method leads to an improvement in diversity. However, we address that this noise injection causes artifact. This can be seen in Figure \ref{fig:only_noise}. Therefore, we propose the structure of the model to remove the noise-induced artifact.
\label{adding_noise}
\begin{figure*}[t]
\begin{center}
   \includegraphics[width=0.8\linewidth]{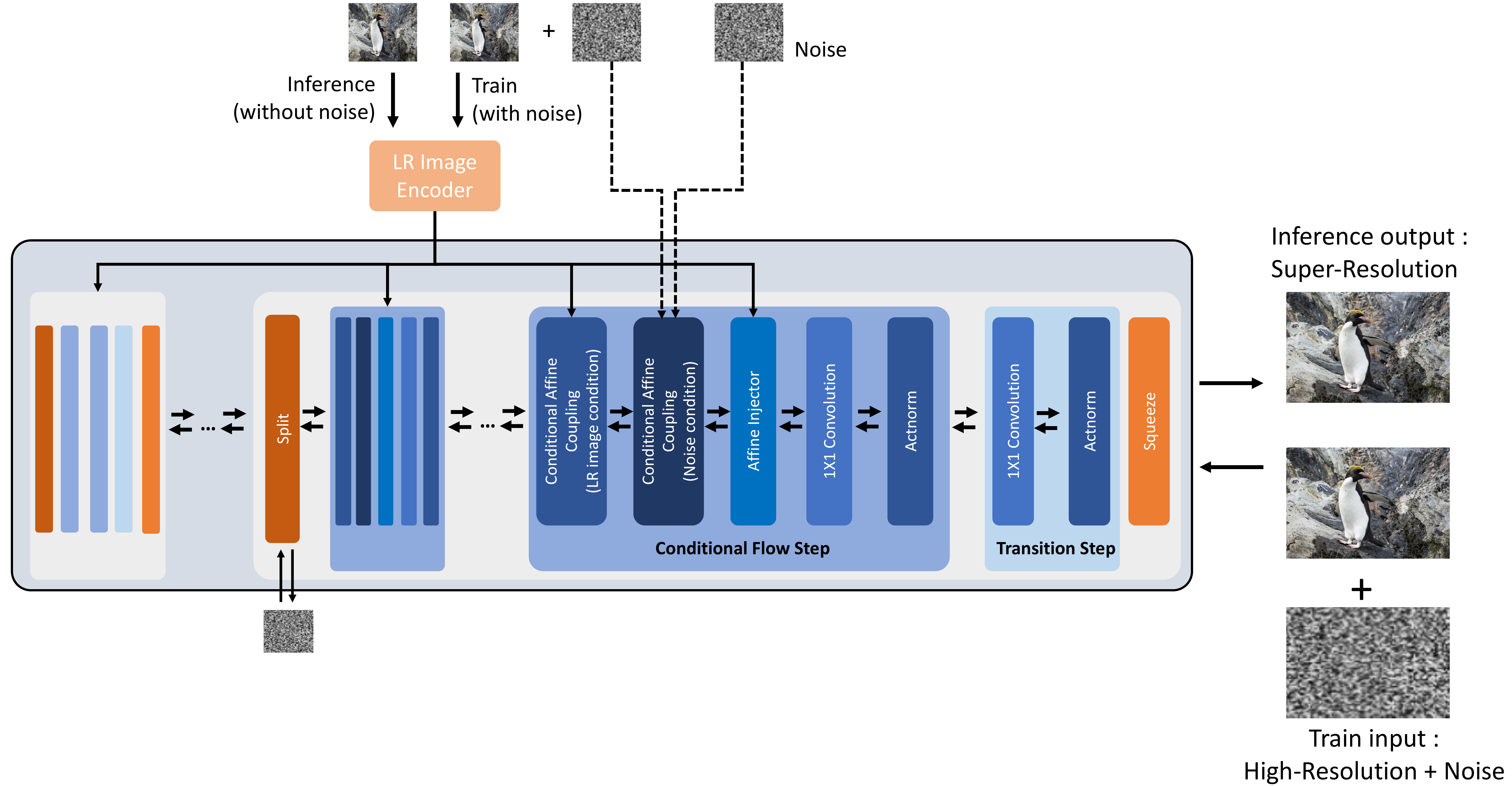}
\end{center}
\captionsetup{justification=raggedright,singlelinecheck=false}
\vspace{-0.3cm}
   \caption{Our method proposes a flow model with noise conditional layer. Our method adds the same noise in LR images and HR images and proceeds with noise conditional training according to these noise distributions.}
\label{fig:overall}
\end{figure*}
\subsection{Noise Conditional Layer}
 It is our motivation to inform the model of information about noise for removing the artifacts caused by adding noise. By training the model with noise information, noise will be reflected for generating images. Due to reflected noise information, images with reduced artifacts can be generated. Therefore, we add layers that inform noise to the model structure. We propose a noise conditional layer.

Our method is as follows. Initially, random value $c$ is obtained from uniform distribution $U(0, M)$ as \cite{kim2020softflow} did. Next, set noise distribution $ N(0, \Sigma)$, where $\Sigma= c^{2}I$. Then, we sample noise vector $v$ from $ N(0, \Sigma)$ and add noise to the original high resolution image $x$ to obtain perturbed data $x^+$. Finally, resize these vector $v$ to get noise vector $w$ for low-resolution images and obtain $y^+$ by adding $w$ to the original low resolution image $y$. 

$$x^+  = x + v $$
$$y^+ = y + w $$
$$f^{-1}(x^+|y^+,v)=z$$

where, $x$ is high resolution image, $y$ is low resolution image, $v$ is noise vector and $w$ is a noise vector which is resized in the same size as $y$. At this point, model $f$ is trained with LR image and noise information. Thus, the goal is to obtain a model conditioned on the noise and LR image that converts latent variable $z$ to $x^+$ given the vector $v$ and  $y^+$. ($i.e.$ $f(z|y^+,v)=x^+$)

Moreover, we conduct noise conditional training in two ways, one for noise itself and one for standard deviation for noise distribution. We proceed both methods in a similar way to the conditional affine coupling of \cite{lugmayr2020srflow}. Although standard deviation conditional training, such as those used in \cite{kim2020softflow}, improves diversity and LPIPS\cite{zhang2018unreasonable}, it tends to create artifact from the generated images. In contrast, with noise conditional training, the numerical performance is slightly lower, but the frequency of occurring artifacts in the generated images is reduced and we finally adopt noise conditional training.

We also deal with the following problem: \begin{itemize}
\item The flow-based model aims to create a complex distribution, $x$, from a simple distribution $z$. However, the manifold data dimension between data $x$ and data $z$ is not always the same. This makes it difficult to predict the distribution of complex data. 
\end{itemize}
Our method solves the mismatch of data distribution. The idea that solves this problem exists in SoftFlow\cite{kim2020softflow}, which adds noise to improve the performance of the flow model. \cite{kim2020softflow} proposes to estimate the conditional distribution of perturbed data to diminish the dimension difference between these data and the target latent variable. The key here is to add noise that is obtained from randomly selected distribution and to use these distribution parameters as conditions. \cite{kim2020softflow} has shown that these methods can experimentally succeed in capturing the innate structure of manifold data. In the same principle, we increase performance for learning the super-resolution space and image visual quality using normalizing flow through adding noise and noise (distribution parameters) conditional training.

\begin{figure}[t]
\begin{center}
   \includegraphics[width=1.0\linewidth]{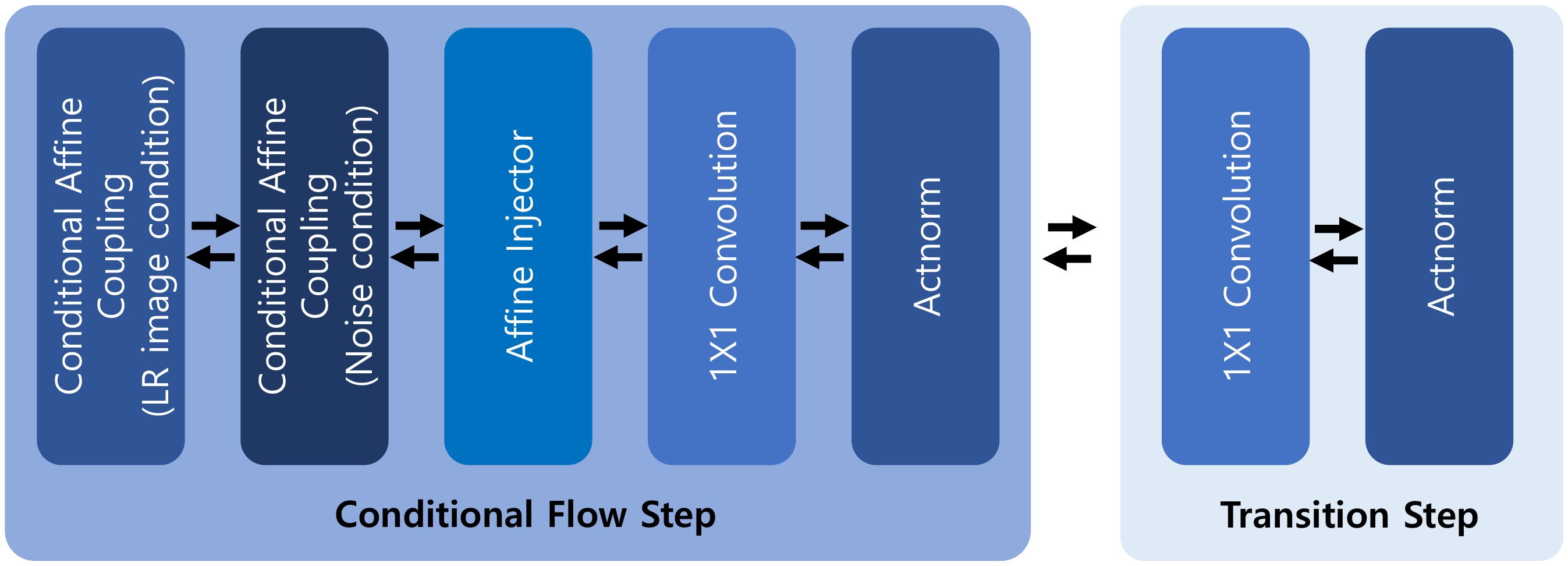}
\end{center}
\captionsetup{justification=raggedright,singlelinecheck=false}
   \vspace{-0.3cm}
   \caption{The key to our model is the Flowstep block in the picture above.}
\label{fig:flowstep}
\vspace{-0.5cm}
\end{figure}

Our model goes through the same process as \cite{lugmayr2020srflow}: squeeze, flowstep, split. Similar to \cite{lugmayr2020srflow}, the LR image is encoded through the low-resolution encoder, which is used for conditional training.  Also, flowstep consists of transition step and conditional flow step, which is equivalent to \cite{lugmayr2020srflow}. The difference between \cite{lugmayr2020srflow} and our model lies at the core of SRFlow, conditional flow step. The noise conditional layer is added to the existing four configuration steps. In other words, it consists of five steps: actnorm, 1$\times$1 convolution, affine injector, and two conditional affine couplings (Noise conditional layer, LR conditional layer). There is the structure of five steps in Figure \ref{fig:flowstep}. Moreover, only negative log-likelihood was used for loss, like \cite{lugmayr2020srflow}. The network is trained with the aim of minimizing the following negative log-likelihood. 
$$-\log{ p_{x|y,v}(x|y,v,\theta)} =$$ $$-\log{p_z(f_\theta(x;y,v))}
-\log{|\det{\frac{\partial{f_\theta}}{\partial{x}}}(x;y,v)|}$$
 where $x$ is high resolution image, $y$ is low resolution image and $v$ is noise vector. During inference, we add a zero vector instead of noise. In inference, because we add zero vector, we need dequantization
 the same as SRFlow. This architecture enables the model to perform noise conditional training, resulting in the reduced artifact.
 
 Noise conditional layer performs better than existing models because our method not only recovers image visual quality dropped by noise injection, but also solves the problem with data distribution mismatch. Noise conditional layer solves the problem of data distribution mismatch and increase the diversity of the data distribution to add noise to the underlying HR image, resulting in performance improvements in both visual quality and diversity. 
\label{noise_conditional_layer}


\begin{figure}[t]
\begin{center}
   \includegraphics[width=1.0\linewidth]{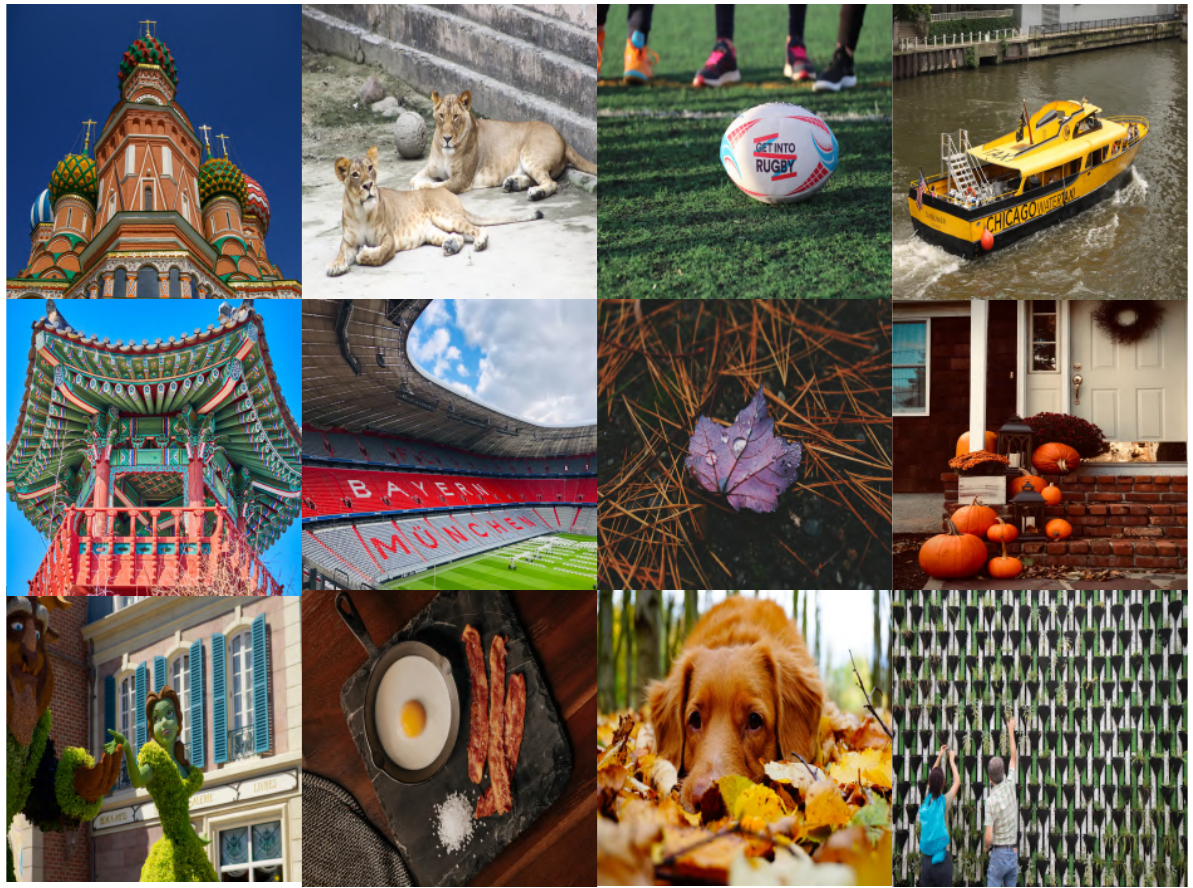}
\end{center}
\vspace{-0.4cm}
   \caption{The sample images of Unsplash 2K}
\label{fig:unsplash2k_sample}
\vspace{-0.4cm}
\end{figure}

\section{Experiments}
\subsection{Dataset and Metric}
We use DF2K dataset, which is a merged dataset with DIV2K~\cite{agustsson2017ntire} and Flick2K\footnote{\url{https://github.com/limbee/NTIRE2017}}, for training the proposed model. DIV2K dataset~\cite{agustsson2017ntire} is composed of 800 train images, 100 validation images, and 100 test images. Similarly, Flick2K contains 2560 training images. Additionally, we use crawled 498 high-resolution images, which are 2K resolution, from the Unsplash website\footnote{\url{https://unsplash.com}} to increase the amount of train data for NTIRE 2021 challenge~\cite{NTIRE2021SRSpace}. We refer to the crawled dataset as Unspalsh2K. Figure \ref{fig:unsplash2k_sample} shows sample images of Unsplash2K dataset. Unsplash2K dataset is publicly available\footnote{\url{https://github.com/dongheehand/unsplash2K}}. For testing, we use the DIV2K validation images because the ground truth images of the DIV2K testset are not publicly available.

We use the following metrics for comparing performance.
\begin{itemize}
  \setlength\itemsep{0.2em}
    \item To compare the visual quality, we use LPIPS~\cite{zhang2018unreasonable} as several works~\cite{ji2020real, lugmayr2020srflow} did. LPIPS~\cite{zhang2018unreasonable} is computed by measuring distance in feature space between two images.
    \item NTIRE 2021 challenge ~\cite{NTIRE2021SRSpace} uses a diversity score to measure the spanning of the super-resolution space. We also use the same metric to evaluate diversity. They calculated the diversity in the following way. Sample 10 images and calculate global best and local best between the samples and the ground truth. The local best is the full image's average of the best score for each pixel out of 10 samples. The global best is the best average of the whole image's score.  The diversity formula is as follows: $$diversity = \cfrac{global\_best - local\_best}{global\_best} * 100$$
    \item To measure the low-resolution consistency, we use the average LR-PSNR of 10 samples. The LR-PSNR is calculated by computing the difference between downsampled prediction image and a low-resolution image.
    \item We use the LR-PSNR worst, which is the minimum value,  of 10 samples to check if artifacts exist. If LR-PSNR worst is low, some of the generated samples have terrible artifacts.
\end{itemize}

\subsection{Implementation Details}
We describe the training details and model hyper-parameters in this section.
For each training step, 18 high-resolution patches are extracted. The size of the extracted patch is 160$\times$160 and extracted patches are used as ground-truth. The low-resolution images downsampled from high-resolution patches via bicubic downsampling are used as input. Both input images and ground-truth images are normalized to $[0, 1]$.

RRDB~\cite{wang2018esrgan} is used for low-resolution encoder for our proposed model. Noise conditional layers are added at the first and the second block based on the order of inference. We use ADAM optimizer ~\cite{kingma2014adam} by setting $\beta_1=0.9$, $\beta_2=0.99$, $\epsilon=10^{-8}$. To augment data, we randomly rotate patches 90, 180, 270 degrees and randomly flip horizontally. The learning rate is initialized to $2\times10^{-4}$. The learning rate is halved at 110K and 165K updates. The other settings are the same as SRFlow \cite{lugmayr2020srflow}. PyTorch is used to implement our model. The code and pretrained models are publicly available\footnote{\url{https://github.com/younggeun-kim/NCSR}}.

\begin{table}[t]
\begin{tabular}{@{}llll@{}}
\toprule
Model            & Diversity & LPIPS  & LR PSNR \\ \midrule
RRDB~\cite{wang2018esrgan}  & 0  & 0.253 & 49.20   \\
ESRGAN~\cite{wang2018esrgan}  & 0  & 0.124 & 39.03   \\
ESRGAN+~\cite{Rakotonirina_2020} & 22.13 & 0.279 & 35.45   \\
SRFlow ~\cite{lugmayr2020srflow} & 25.26     & 0.120 & 49.97   \\
\textbf{NCSR (Ours)}   & \textcolor{blue}{26.72}     & \textcolor{blue}{0.119} & \textcolor{blue}{50.75}   \\
\textbf{NCSR* (Ours)} & \textcolor{red}{26.79}    & \textcolor{red}{0.118} & \textcolor{red}{50.88}   \\ \bottomrule
\end{tabular}
\captionsetup{justification=raggedright,singlelinecheck=false}
\vspace{-0.3cm}
\caption{General image SR $\times$4 results on the 100 validation images of the DIV2K dataset}
\label{tab:X4_results}
\end{table}

\subsection{Comparision with other models}
To show superiority of our model, we compare our method with other super-resolution methods. We compare performance with RRDB~\cite{wang2018esrgan}, ESRGAN~\cite{wang2018esrgan}, ESRGAN+~\cite{Rakotonirina_2020} and SRFlow~\cite{lugmayr2020srflow}. RRDB~\cite{wang2018esrgan} is PSNR oriented model which is trained with $L_1$ loss. ESRGAN~\cite{wang2018esrgan} and ESRGAN+~\cite{Rakotonirina_2020} are GAN based model. SRFlow~\cite{lugmayr2020srflow} is Flow-based model.

We evaluated the performance of these models with three metrics: LPIPS, diversity score, LR-PSNR. For $\times$4 SR model, our proposed model outperformed the state-of-the-art model SRFlow on all metrics as shown in table \ref{tab:X4_results}. Furthermore, our model achieves superior result than GAN based model in terms of LPIPS, which is a perceptual measure.

Similarly, we measured the performance of our model for $\times$8 SR. We do not compare with ESRGAN+, because there is no model for $\times$8 SR. In $\times$8 task, our method shows a better diversity score than SRFlow.  Comparable results are also shown in terms of LPIPS and LR-PSNR. 

We also do an experiment by adding Unsplash2K, which is an extra training dataset. When we add Unsplash2K to train data, all metrics are slightly better for both $\times$4 SR and $\times$8 SR. In table \ref{tab:X4_results} and \ref{tab:X8_results}, NCSR means the model trained with DF2K only, and NCSR* means the model trained with DF2K and Unsplash2K.

\begin{figure*}[t]
		\captionsetup[subfloat]{labelformat=empty}
		\begin{center}
			\newcommand{\rowArg}{2.46cm}
			\newcommand{\fullSize}{5.26cm}
			\newcommand{\fullHalf}{2.78cm}
			\newcommand{\patchSize}{2.3cm}
			\scriptsize
			\setlength\tabcolsep{0.1cm}
			\begin{tabular}[b]{c c}
				\begin{tabular}[b]{c c c}
					\multirow{2}{*}[\rowArg]{
						\subfloat[0850 from DIV2K]
						{\includegraphics[width = \fullHalf, height = \fullSize]	{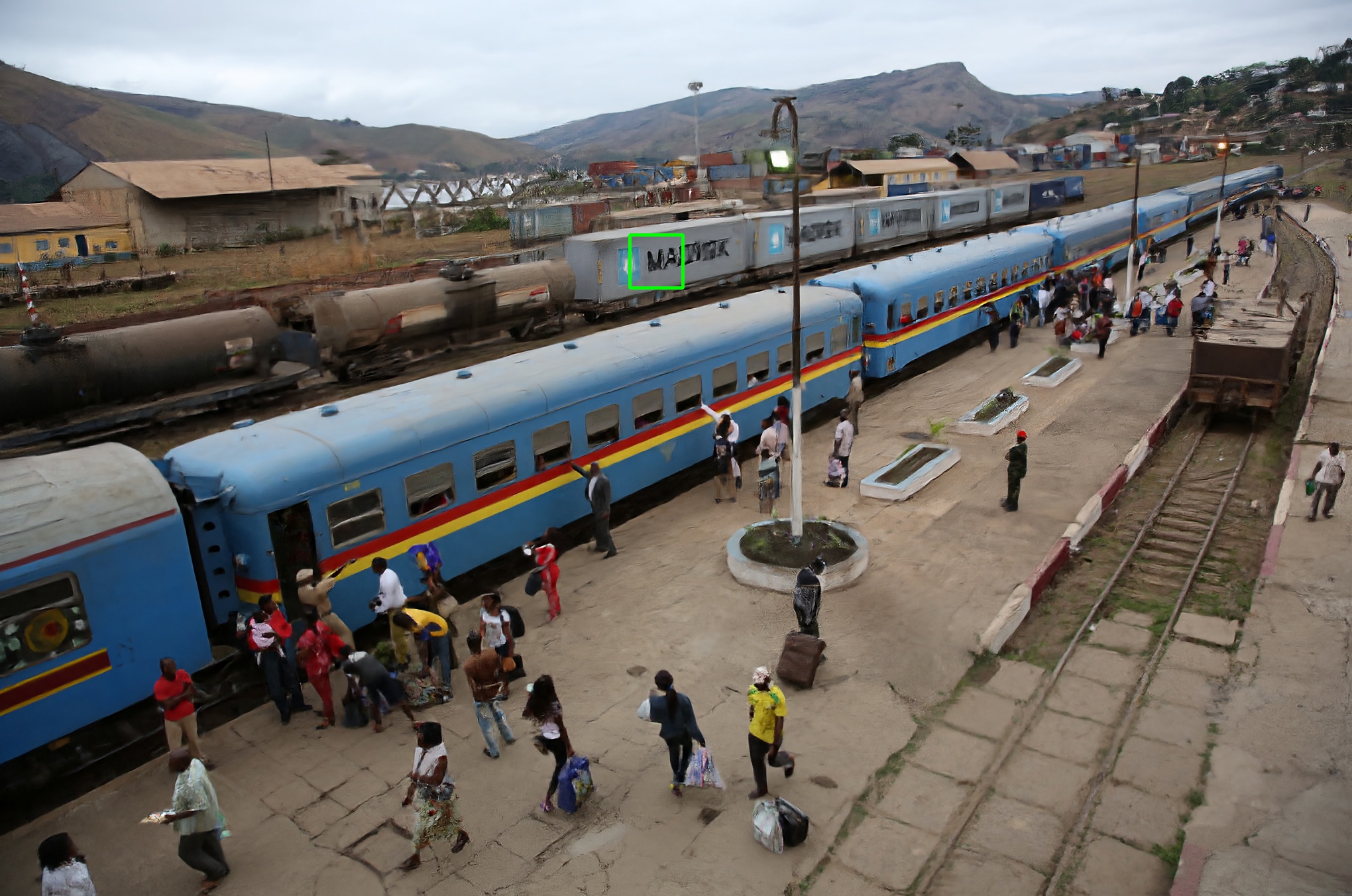}}} &
					\subfloat[LR \protect\linebreak]
					{\includegraphics[width = \patchSize, height = \patchSize]	{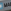}} &
					\subfloat[RRDB \protect\linebreak]
					{\includegraphics[width = \patchSize, height = \patchSize]
						{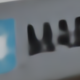}} \\ [-0.5cm] &
					\subfloat[{SRFlow} \protect\linebreak]
					{\includegraphics[width = \patchSize, height = \patchSize]	{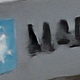}} &
					\subfloat[\textbf{NCSR (Ours)} \protect\linebreak]
					{\includegraphics[width = \patchSize, height = \patchSize]	{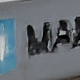}} 
				\end{tabular}
				&  
				\begin{tabular}[b]{c c c}
					\multirow{2}{*}[\rowArg]{
						\subfloat[0847 from DIV2K]
						{\includegraphics[width = \fullHalf, height = \fullSize]
						{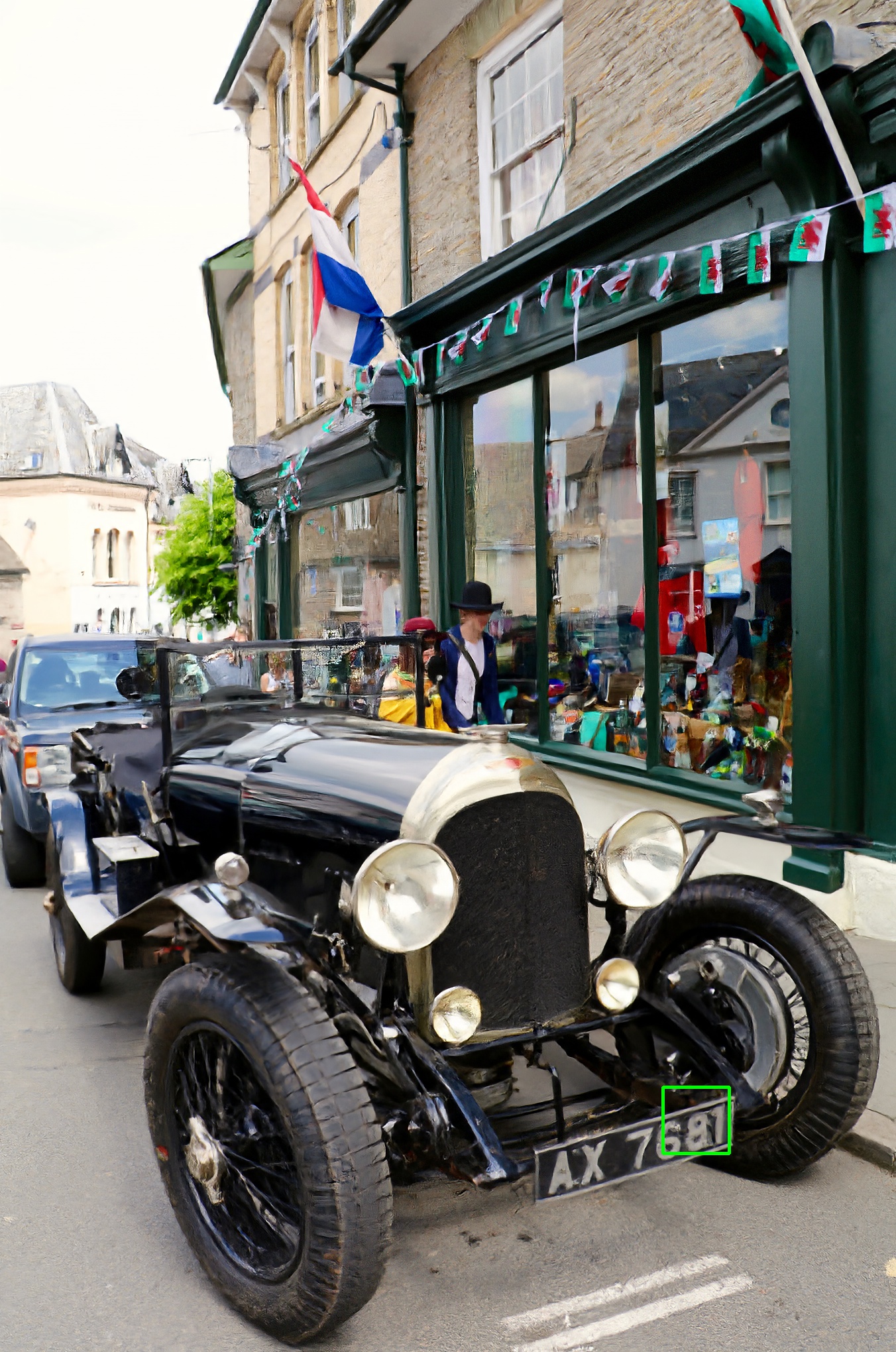}}} &
					\subfloat[LR \protect\linebreak]
					{\includegraphics[width = \patchSize, height = \patchSize]	{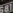}} &
					\subfloat[RRDB \protect\linebreak]
					{\includegraphics[width = \patchSize, height = \patchSize]	{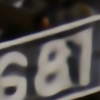}} \\ [-0.5cm] &
					\subfloat[SRFlow \protect\linebreak]
					{\includegraphics[width = \patchSize, height = \patchSize]	{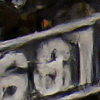}} &
					\subfloat[\textbf{NCSR (Ours)} \protect\linebreak]
					{\includegraphics[width = \patchSize, height = \patchSize]	{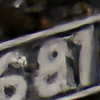}} 
				\end{tabular}
			\end{tabular}
		\end{center}
		
		\vspace{-1.0cm}
		\caption{Qualitative comparisons with other methods for $\times$8 SR model.}
		
		\label{fig_qual_results}
	\end{figure*}

\begin{figure*}[t]
\begin{center}

\newcommand{\size}{0.19}
\includegraphics[width=\size\linewidth]{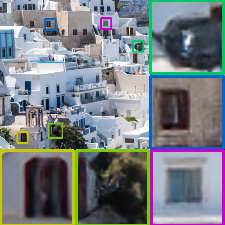}
\includegraphics[width=\size\linewidth]{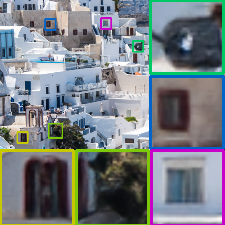}
\includegraphics[width=\size\linewidth]{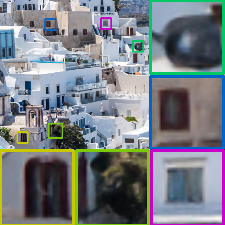}
\includegraphics[width=\size\linewidth]{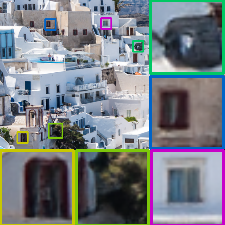}
\includegraphics[width=\size\linewidth]{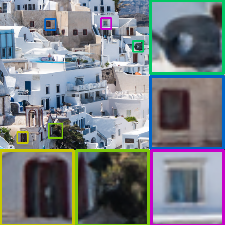}
\includegraphics[width=\size\linewidth]{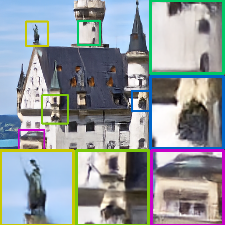}
\includegraphics[width=\size\linewidth]{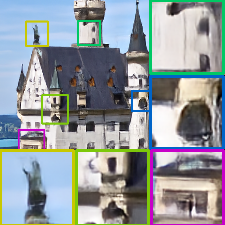}
\includegraphics[width=\size\linewidth]{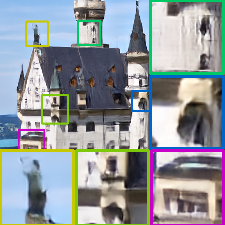}
\includegraphics[width=\size\linewidth]{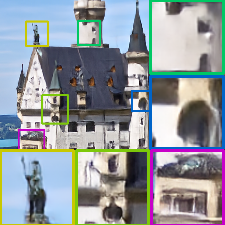}
\includegraphics[width=\size\linewidth]{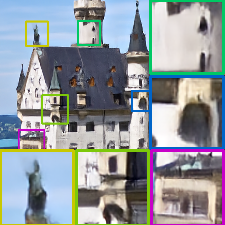}
\end{center}
\vspace{-5mm}
\caption{Random samples generated by NCSR. Upper : $\times$4 SR model, Lower : $\times$8 SR model}
\vspace{-3mm}
\label{fig:diversity_div2k}
\end{figure*}

Qualitative results are shown in Figure \ref{fig_qual_results}, \ref{fig:diversity_div2k}. Figure \ref{fig_qual_results} shows that our proposed model reconstruct textures and details compared to other works. Random samples generated by our model are shown in Figure \ref{fig:diversity_div2k}.

For the flow-based SR model, we set the temperature as 0.9. However, the temperature is 0.85 for our $\times$8 SR model trained with DF2K only. 

\begin{table}[t]
\begin{tabular}{@{}llll@{}}
\toprule
Model            & Diversity & LPIPS  & LR PSNR \\ \midrule
RRDB~\cite{wang2018esrgan} & 0         & 0.419  & 45.43   \\
ESRGAN~\cite{wang2018esrgan} & 0         & 0.277  & 31.35   \\
SRFlow~\cite{lugmayr2020srflow} & 25.31     & \textcolor{blue}{0.272} & \textcolor{red}{50.00}   \\
\textbf{NCSR (Ours)}             &\textcolor{red} {26.8}      & 0.278 & 44.55   \\
\textbf{NCSR* (Ours)} & \textcolor{blue}{25.7}      & \textcolor{red}{0.253} & \textcolor{blue}{49.97} \\ \bottomrule
\end{tabular}
\captionsetup{justification=raggedright,singlelinecheck=false}
\vspace{-0.3cm}
\caption{General image SR $\times$8results on the 100 validation images of the DIV2K dataset}
\label{tab:X8_results}
\end{table}

\subsection{Ablation Study}
In this section, we analyze the performance of the proposed model according to three factors.

\paragraph{
Std conditional layer vs. noise conditional layer
} We compare the performance of which value is used for conditional layer. While the noise conditional layer uses sampled noise, the standard deviation conditional layer uses the standard deviation of sampled noise, which is injected into high-resolution images.
As you can see in the table \ref{tab:lrpsnr}, the standard deviation conditional layer makes LR-PSNR worst value lower than our noise conditional layer. This means that the standard deviation conditional layer creates many artifacts. It is because the standard deviation conditional layer did not provide sufficient information to the model. Therefore, the standard deviation is not enough for noise-aware training when noise-injected inputs are given. 
\begin{table}[ht]
\begin{tabular}{llll}
\toprule
{} &  LPIPS & LR PSNR & Div. Score \\
Team            &        &         &            \\
\midrule
svnit\_ntnu      &  0.355 &   27.52 &      1.871 \\
SYSU-FVL        &  0.244 &   49.33 &      8.735 \\
nanbeihuishi    &  0.161 &   50.46 &     12.447 \\
SSS             &  0.110 &   44.70 &     13.285 \\
FudanZmic21     &  0.273 &   47.20 &     16.450 \\
FutureReference &  0.165 &   37.51 &     19.636 \\
SR\_DL           &  0.234 &   39.80 &     20.508 \\
CIPLAB          &  0.121 &   50.70 &     23.091 \\
BeWater         &  0.137 &   49.59 &     23.948 \\
njtech\&seu      &  0.149 &   46.74 &     26.924 \\
\textbf{Deepest(ours)}         &  \textbf{0.117} &   \textbf{50.54} &     \textbf{26.041} \\
\bottomrule
\end{tabular}
\vspace{-0.1cm}
\caption{\textbf{Quantitative results for NTIRE 2021 Challenge on Learning Super Resolution Space on $\times$4 track} 
 }
 \label{tab:Result_NTIRE_X4}
 \end{table}

\begin{table}[ht]
\begin{tabular}{llll}
\toprule
{} &  LPIPS & LR PSNR & Div. Score \\
Team            &        &         &            \\
\midrule
svnit\_ntnu      &  0.481 &   25.55 &      4.516 \\
SYSU-FVL        &  0.415 &   47.27 &      8.778 \\
SSS             &  0.237 &   37.43 &     13.548 \\
FudanZmic21     &  0.496 &   46.78 &     14.287 \\
SR\_DL           &  0.311 &   42.28 &     14.817 \\
FutureReference &  0.291 &   36.51 &     17.985 \\
CIPLAB          &  0.266 &   50.86 &     23.320 \\
BeWater         &  0.297 &   49.63 &     23.700 \\
njtech\&seu      &  0.366 &   29.65 &     28.193 \\
\textbf{Deepest(ours)}        &  \textbf{0.259} &   \textbf{48.64} &     \textbf{26.941} \\
\bottomrule
\end{tabular}
\vspace{-0.1cm}
\caption{\textbf{Quantitative results for NTIRE 2021 Challenge on Learning Super Resolution Space on $\times$8 track} 
 }
 \label{tab:Result_NTIRE_X8}
 \end{table} 
\paragraph{
With or without noise conditional layer
}
We investigate the effect of the noise conditional layer. Without noise conditional layer, noises are injected in input images. As you can see in Table \ref{tab:noise_compare}, LPIPS is high without a noise conditional layer. In other words, the image visual quality is not good. Furthermore, LR-PSNR is low, which means that there are lots of artifacts. Images with artifacts can be seen in Figure \ref{fig:only_noise}. Table \ref{tab:noise_compare} shows the detailed performance comparison with the presence of noise conditional layer.

\paragraph{
Where to add noise conditional layer
}
We also experimentally show that it is recommended that the noise conditional layer is only included in the first block and the second block based on the order of inference. The model with a noise conditional layer in all blocks generates images with artifact. They show slightly better scores in terms of diversity, but due to the occurrence of these artifacts, we adopt to add our noise conditional layer in the first and second block. When generating super-resolution output, we need an interval at the end of network to generate the image without such noise because of noise added in the input and the noise conditional layer. 

That is to say, we find that there should be noise-free block at the end of network. The noise-free block is the block that there is no noise conditional layer. Therefore, if there is no \textbf{noise-free block}, this could be the factor that make the artifact. You can show the results described above in the following table \ref{tab:lrpsnr}.

\begin{table}[t]
\begin{tabular}{@{}lll@{}}
\toprule
Model         & w/o NCL & with NCL   \\ \midrule
Diversity     & 25.38      & 26.72  \\
LPIPS         & 0.1228     & 0.1193 \\ 
LR PSNR       & 50.08      & 50.75  \\
LR PSNR-worst & 47.32      & 49.14  \\ \bottomrule
\end{tabular}
\captionsetup{justification=raggedright,singlelinecheck=false}
\caption{Performance comparison between model with noise conditional layer and without noise conditional layer
 }
\label{tab:noise_compare}
\end{table}

\section{NTIRE2021 challenge}

Our method, NCSR, scored high in both tracks of NTIRE 2021 Learning Super Resolution Space Challenge ~\cite{NTIRE2021SRSpace}. To measure how much information is preserved in the super-resolution image from the low-resolution image, the competition measured the LR-PSNR. In this competition, a team with high perceptual image visual quality and diversity scores becomes the winner in the case that LR-PSNR exceeds only 45. Among teams with LR-PSNR over 45, we take the first place in LPIPS and the second place in diversity score for $\times$4 track. Moreover, in the $\times$8 track, our model takes the first place in both LPIPS and diversity scores. The quantitative results of NTIRE 2021 Super-Resolution Challenge are shown in Table \ref{tab:Result_NTIRE_X4}, Table \ref{tab:Result_NTIRE_X8}.

\section{Conclusion}
We propose noise conditioned flow model for learning super-resolution space. Our proposed model uses a noise conditional layer to generate more diverse super-resolution images with high visual quality. 

To learn more diverse data distribution, we add a random noise to images. Although data distribution is broader, adding noise cause artifacts with terrible quality in super-resolution images. Therefore, a noise conditional layer is proposed for stable training when noise is added to images. By using the noise conditional layer, we can obtain more diverse super-resolution images without visual degradation. We show the superiority of our proposed model on the DIV2K dataset under several settings. Furthermore, our proposed model achieves high quantitative results on NTIRE 2021 Super-Resolution Challenge~\cite{NTIRE2021SRSpace}.

\begin{table}[tbp]
\scriptsize

\begin{center}
\begin{tabular}{|c|c|c|c|c|c|}
\hline
Noise Conditional Layer & \xmark & \xmark & \checkmark &\checkmark & \checkmark
\\ %
Std Conditional Layer & \xmark & \checkmark & \xmark&\xmark & \xmark
\\
Noise-free block & \xmark & \checkmark & \xmark&\checkmark & \checkmark
\\
Add extra data & \xmark & \xmark & \xmark&\xmark & \checkmark
\\
\hline
\hline
LR-PSNR worst & 47.32 & 45.78 &49.01 & 49.14 & 50.13 
\\
\hline
\end{tabular}
\end{center}
\vspace{-3mm}

\captionsetup{justification=raggedright,singlelinecheck=false}
\caption{LR-PSNR worst comparison for Ablation Study}
\label{tab:lrpsnr}
\end{table}

{\small
\bibliographystyle{ieee_fullname}
\bibliography{egbib}

\begin{thebibliography}{10}\itemsep=-1pt

\bibitem{agustsson2017ntire}
Eirikur Agustsson and Radu Timofte.
\newblock Ntire 2017 challenge on single image super-resolution: Dataset and
  study.
\newblock In {\em Proceedings of the IEEE Conference on Computer Vision and
  Pattern Recognition Workshops}, pages 126--135, 2017.

\bibitem{ardizzone2019guided}
Lynton Ardizzone, Carsten L{\"u}th, Jakob Kruse, Carsten Rother, and Ullrich
  K{\"o}the.
\newblock Guided image generation with conditional invertible neural networks.
\newblock {\em arXiv preprint arXiv:1907.02392}, 2019.

\bibitem{chen2018efficient}
Yuhua Chen, Feng Shi, Anthony~G Christodoulou, Yibin Xie, Zhengwei Zhou, and
  Debiao Li.
\newblock Efficient and accurate mri super-resolution using a generative
  adversarial network and 3d multi-level densely connected network.
\newblock In {\em International Conference on Medical Image Computing and
  Computer-Assisted Intervention}, pages 91--99. Springer, 2018.

\bibitem{dinh2014nice}
Laurent Dinh, David Krueger, and Yoshua Bengio.
\newblock Nice: Non-linear independent components estimation.
\newblock {\em arXiv preprint arXiv:1410.8516}, 2014.

\bibitem{dinh2016density}
Laurent Dinh, Jascha Sohl-Dickstein, and Samy Bengio.
\newblock Density estimation using real nvp.
\newblock {\em arXiv preprint arXiv:1605.08803}, 2016.

\bibitem{dong2015image}
Chao Dong, Chen~Change Loy, Kaiming He, and Xiaoou Tang.
\newblock Image super-resolution using deep convolutional networks.
\newblock {\em IEEE transactions on pattern analysis and machine intelligence},
  38(2):295--307, 2015.

\bibitem{goodfellow2014generative}
Ian~J Goodfellow, Jean Pouget-Abadie, Mehdi Mirza, Bing Xu, David Warde-Farley,
  Sherjil Ozair, Aaron Courville, and Yoshua Bengio.
\newblock Generative adversarial networks.
\newblock {\em arXiv preprint arXiv:1406.2661}, 2014.

\bibitem{grathwohl2018ffjord}
Will Grathwohl, Ricky~TQ Chen, Jesse Bettencourt, Ilya Sutskever, and David
  Duvenaud.
\newblock Ffjord: Free-form continuous dynamics for scalable reversible
  generative models.
\newblock {\em arXiv preprint arXiv:1810.01367}, 2018.

\bibitem{he2016deep}
Kaiming He, Xiangyu Zhang, Shaoqing Ren, and Jian Sun.
\newblock Deep residual learning for image recognition.
\newblock In {\em Proceedings of the IEEE conference on computer vision and
  pattern recognition}, pages 770--778, 2016.

\bibitem{hu2018squeeze}
Jie Hu, Li Shen, and Gang Sun.
\newblock Squeeze-and-excitation networks.
\newblock In {\em Proceedings of the IEEE conference on computer vision and
  pattern recognition}, pages 7132--7141, 2018.

\bibitem{huang2017densely}
Gao Huang, Zhuang Liu, Laurens Van Der~Maaten, and Kilian~Q Weinberger.
\newblock Densely connected convolutional networks.
\newblock In {\em Proceedings of the IEEE conference on computer vision and
  pattern recognition}, pages 4700--4708, 2017.

\bibitem{ji2020real}
Xiaozhong Ji, Yun Cao, Ying Tai, Chengjie Wang, Jilin Li, and Feiyue Huang.
\newblock Real-world super-resolution via kernel estimation and noise
  injection.
\newblock In {\em Proceedings of the IEEE/CVF Conference on Computer Vision and
  Pattern Recognition Workshops}, pages 466--467, 2020.

\bibitem{johnson2016perceptual}
Justin Johnson, Alexandre Alahi, and Li Fei-Fei.
\newblock Perceptual losses for real-time style transfer and super-resolution.
\newblock In {\em European conference on computer vision}, pages 694--711.
  Springer, 2016.

\bibitem{kim2020softflow}
Hyeongju Kim, Hyeonseung Lee, Woo~Hyun Kang, Joun~Yeop Lee, and Nam~Soo Kim.
\newblock Softflow: Probabilistic framework for normalizing flow on manifolds.
\newblock {\em arXiv preprint arXiv:2006.04604}, 2020.

\bibitem{kim2016accurate}
Jiwon Kim, Jung~Kwon Lee, and Kyoung~Mu Lee.
\newblock Accurate image super-resolution using very deep convolutional
  networks.
\newblock In {\em Proceedings of the IEEE conference on computer vision and
  pattern recognition}, pages 1646--1654, 2016.

\bibitem{kingma2014adam}
Diederik~P Kingma and Jimmy Ba.
\newblock Adam: A method for stochastic optimization.
\newblock {\em arXiv preprint arXiv:1412.6980}, 2014.

\bibitem{kingma2018glow}
Diederik~P Kingma and Prafulla Dhariwal.
\newblock Glow: Generative flow with invertible 1x1 convolutions.
\newblock {\em arXiv preprint arXiv:1807.03039}, 2018.

\bibitem{ledig2017photo}
Christian Ledig, Lucas Theis, Ferenc Husz{\'a}r, Jose Caballero, Andrew
  Cunningham, Alejandro Acosta, Andrew Aitken, Alykhan Tejani, Johannes Totz,
  Zehan Wang, et~al.
\newblock Photo-realistic single image super-resolution using a generative
  adversarial network.
\newblock In {\em Proceedings of the IEEE conference on computer vision and
  pattern recognition}, pages 4681--4690, 2017.

\bibitem{li2018super}
Zhan Li, Qingyu Peng, Bir Bhanu, Qingfeng Zhang, and Haifeng He.
\newblock Super resolution for astronomical observations.
\newblock {\em Astrophysics and Space Science}, 363(5):1--15, 2018.

\bibitem{lim2017enhanced}
Bee Lim, Sanghyun Son, Heewon Kim, Seungjun Nah, and Kyoung Mu~Lee.
\newblock Enhanced deep residual networks for single image super-resolution.
\newblock In {\em Proceedings of the IEEE conference on computer vision and
  pattern recognition workshops}, pages 136--144, 2017.

\bibitem{NTIRE2021SRSpace}
Andreas Lugmayr, Martin Danelljan, Radu Timofte, et~al.
\newblock Ntire 2021 learning the super-resolution space challenge: Methods and
  results.
\newblock {\em CVPR Workshops}, 2021.

\bibitem{lugmayr2020srflow}
Andreas Lugmayr, Martin Danelljan, Luc Van~Gool, and Radu Timofte.
\newblock Srflow: Learning the super-resolution space with normalizing flow.
\newblock In {\em European Conference on Computer Vision}, pages 715--732.
  Springer, 2020.

\bibitem{noh2019better}
Junhyug Noh, Wonho Bae, Wonhee Lee, Jinhwan Seo, and Gunhee Kim.
\newblock Better to follow, follow to be better: Towards precise supervision of
  feature super-resolution for small object detection.
\newblock In {\em Proceedings of the IEEE/CVF International Conference on
  Computer Vision}, pages 9725--9734, 2019.

\bibitem{puschmann2005super}
Klaus~G Puschmann and Franz Kneer.
\newblock On super-resolution in astronomical imaging.
\newblock {\em Astronomy \& Astrophysics}, 436(1):373--378, 2005.

\bibitem{radford2015unsupervised}
Alec Radford, Luke Metz, and Soumith Chintala.
\newblock Unsupervised representation learning with deep convolutional
  generative adversarial networks.
\newblock {\em arXiv preprint arXiv:1511.06434}, 2015.

\bibitem{Rakotonirina_2020}
Nathanael~Carraz Rakotonirina and Andry Rasoanaivo.
\newblock Esrgan+: Further improving enhanced super-resolution generative
  adversarial network.
\newblock {\em ICASSP 2020 - 2020 IEEE International Conference on Acoustics,
  Speech and Signal Processing (ICASSP)}, May 2020.

\bibitem{shi2013cardiac}
Wenzhe Shi, Jose Caballero, Christian Ledig, Xiahai Zhuang, Wenjia Bai, Kanwal
  Bhatia, Antonio M Simoes~Monteiro de Marvao, Tim Dawes, Declan O’Regan, and
  Daniel Rueckert.
\newblock Cardiac image super-resolution with global correspondence using
  multi-atlas patchmatch.
\newblock In {\em International conference on medical image computing and
  computer-assisted intervention}, pages 9--16. Springer, 2013.

\bibitem{wang2018esrgan}
Xintao Wang, Ke Yu, Shixiang Wu, Jinjin Gu, Yihao Liu, Chao Dong, Yu Qiao, and
  Chen Change~Loy.
\newblock Esrgan: Enhanced super-resolution generative adversarial networks.
\newblock In {\em Proceedings of the European Conference on Computer Vision
  (ECCV) Workshops}, pages 0--0, 2018.

\bibitem{winkler2019learning}
Christina Winkler, Daniel Worrall, Emiel Hoogeboom, and Max Welling.
\newblock Learning likelihoods with conditional normalizing flows.
\newblock {\em arXiv preprint arXiv:1912.00042}, 2019.

\bibitem{zhang2018unreasonable}
Richard Zhang, Phillip Isola, Alexei~A. Efros, Eli Shechtman, and Oliver Wang.
\newblock The unreasonable effectiveness of deep features as a perceptual
  metric, 2018.

\bibitem{zhang2018image}
Yulun Zhang, Kunpeng Li, Kai Li, Lichen Wang, Bineng Zhong, and Yun Fu.
\newblock Image super-resolution using very deep residual channel attention
  networks.
\newblock In {\em Proceedings of the European conference on computer vision
  (ECCV)}, pages 286--301, 2018.

\bibitem{zhang2018residual}
Yulun Zhang, Yapeng Tian, Yu Kong, Bineng Zhong, and Yun Fu.
\newblock Residual dense network for image super-resolution.
\newblock In {\em Proceedings of the IEEE conference on computer vision and
  pattern recognition}, pages 2472--2481, 2018.

\bibitem{zou2011very}
Wilman~WW Zou and Pong~C Yuen.
\newblock Very low resolution face recognition problem.
\newblock {\em IEEE Transactions on image processing}, 21(1):327--340, 2011.

\end{thebibliography}
}

\end{document}